\documentclass[sigconf]{acmart}

\AtBeginDocument{%
  \providecommand\BibTeX{{%
    \normalfont B\kern-0.5em{\scshape i\kern-0.25em b}\kern-0.8em\TeX}}}

\usepackage{amsmath}
\newtheorem{definition}{Definition}
\newtheorem{theorem}{Theorem}

\newcommand\newcite[1]{\citeauthor{#1}~\cite{#1}}
\newcommand{\thickhline}{\noalign{\hrule height 1pt}}
\usepackage{enumitem}
\usepackage{multirow}
\usepackage{multicol}
\usepackage{mathrsfs}
\usepackage{algorithmic}
\usepackage[ruled]{algorithm2e}

\copyrightyear{2022} 
\acmYear{2022} 
\setcopyright{acmcopyright}\acmConference[WWW '22]{Proceedings of the ACM Web Conference 2022}{April 25--29, 2022}{Virtual Event, Lyon, France}
\acmBooktitle{Proceedings of the ACM Web Conference 2022 (WWW '22), April 25--29, 2022, Virtual Event, Lyon, France}
\acmPrice{15.00}
\acmDOI{10.1145/3485447.3512096}
\acmISBN{978-1-4503-9096-5/22/04}

\begin{document}
\fancyhead{}

\title{MBCT: Tree-Based Feature-Aware Binning for Individual Uncertainty Calibration}

\author{Siguang Huang$^1$, Yunli Wang$^1$, Lili Mou$^2$, Huayue Zhang$^1$, Han Zhu$^1$, Chuan Yu$^1$, Bo Zheng$^1$}
\renewcommand{\shortauthors}{
}
\affiliation{
\institution{$^1$Alibaba Group, China\\
$^2$Dept. Computing Science, Alberta Machine Intelligence Institute (Amii), University of Alberta, Canada}
\and
\{siguang.hsg, ruoyu.wyl, huayue.zhy, zhuhan.zh, yuchuan.yc, bozheng\}@alibaba-inc.com
\and
doublepower.mou@gmail.com
\country{}
}

%% The abstract is a short summary of the work to be presented in the
%% article.
\begin{abstract}
  Most machine learning classifiers only concern classification accuracy, while certain applications (such as medical diagnosis, meteorological forecasting, and computation advertising) require the model to predict the true probability, known as a calibrated estimate. 
   In previous work, researchers have developed several calibration methods to post-process the outputs of a predictor to obtain calibrated values, such as binning and scaling methods. Compared with scaling, binning methods are shown to have distribution-free theoretical guarantees, which motivates us to prefer binning methods for calibration. However, we notice that existing binning methods have several drawbacks: (a) the binning scheme only considers the original prediction values, thus limiting the calibration performance; and (b) the binning approach is non-individual, mapping multiple samples in a bin to the same value, and thus is not suitable for order-sensitive applications. In this paper, we propose a feature-aware binning framework, called Multiple Boosting Calibration Trees (MBCT), along with a multi-view calibration loss to tackle the above issues. Our MBCT optimizes the binning scheme by the tree structures of features, and adopts a linear function in a tree node to achieve individual calibration. Our MBCT is non-monotonic, and has the potential to improve order accuracy, due to its learnable binning scheme and the individual calibration. 
  We conduct comprehensive experiments on three datasets in different fields. Results show that our method outperforms all competing models in terms of both calibration error and order accuracy.
  We also conduct simulation experiments, justifying that the proposed multi-view calibration loss is a better metric in modeling calibration error. In addition, our approach is deployed in a real-world online advertising platform; an A/B test over two weeks further demonstrates the effectiveness and great business value of our approach.
\end{abstract}

%% The code below is generated by the tool at http://dl.acm.org/ccs.cfm.


%% Keywords. The author(s) should pick words that accurately describe
%% the work being presented. Separate the keywords with commas.

%% This command processes the author and affiliation and title
%% information and builds the first part of the formatted document.
\maketitle

\section{Introduction}
In recent years, machine learning is widely used for classification in various fields. A large number of tasks not only concern the classification accuracy, but also expect that the output of the machine learning model reflects the true frequency of an event, such as medical diagnosis \cite{medical_jiang2012calibrating}, meteorological forecasting \cite{weather_brocker2009reliability,weather_degroot1983comparison,weather_gneiting2005weather}, and computation advertising \cite{mcb_bid,bcb_bid}. Take computation advertising as an example. If the click-through rate on 1000 page views is 0.05, we expect approximately 50 clicks. Unfortunately, the prediction of most machine learning models (e.g., logistic regression and deep neural networks) usually deviates from the true probability, either overestimated or underestimated. This in turn leads to the lack of robustness and interpretability for these models, limiting their application in safe-critical tasks. 

To mitigate the estimation deviation, researchers have developed various methods to better estimate the true probability. Several studies directly learn a calibrated predictor \cite{individual_zhao2020}, whereas others adopt a post-hoc calibrator for the predictor \cite{temperature_scaling,verified,mix_n_match}. The post-hoc methods, which are commonly known as \textit{calibration}, are widely used, as they can be applied to off-the-shelf predictors. Existing post-hoc calibration can be further divided into three groups: parametric (e.g., scaling), non-parametric (e.g., binning), and their hybrid. Scaling methods adopt a parametric function with a certain characteristic assumption, which is usually data efficient, but lacks theoretical guarantees \cite{verified,distribution_free}. Binning methods divide the data samples into finite buckets (bins), and then calibrate the samples in the bucket by its average prediction, which provides distribution-free theoretical guarantees but requires more data samples. \newcite{verified} propose the scaling-binning method that combines Platt scaling \cite{platt1999probabilistic} and histogram binning \cite{uniform_mass} to balance the data efficiency and distribution-free theoretical guarantees. In our work, we also consider the hybrid of binning and scaling.

Although the widely used binning methods do not have distribution assumptions, we find the binning scheme of previous methods is too simple, limiting the calibration performance. Here, the binning scheme is a way to divide data samples into a finite number of bins. For example, histogram binning with the uniform-mass scheme \cite{uniform_mass} sorts the samples by the predictions and then splits the samples into multiple bins, in which the number of samples are approximately the same. Histogram binning further assumes samples with close predictions have the same true probability, which is a widely used binning approach. However, even though the average prediction probability is close to the ground-truth, some samples in a bin may be overestimated, whereas others may be underestimated, because the samples in the same bin may still exhibit different bias patterns. We observe that the overestimation and underestimation are related to both the data sample itself and the predictor outputs. Thus, we should consider both the input features and predicted output into consideration for calibration, and put samples sharing a similar bias pattern into the same bin.

To this end, we propose a feature-aware binning framework that learns an improved binning scheme for calibration, which takes both input features and the output prediction of a predictor as the calibrator's inputs. We further employ multiple boosting trees as the backbone model for feature-aware binning due to the following considerations: 1) a path from the tree root to the leaf is a natural bin, which largely improves model interpretation; 2) tree-based methods enjoy the flexibility of handling various loss functions for node split; and 3) boosting combines multiple weak tree learners to a strong learner for better performance.
We name the proposed framework as Multiple Boosting Calibration Trees (MBCT).

Moreover, we notice that traditional evaluation metrics usually adopt the uniform-mass scheme with sorted samples, which only reflects the calibration error on a certain group of partitions. Ideally, the loss of a well-performing calibrator should be near 0 under any partition. Therefore, we propose a multi-view calibration error (MVCE) as both the node-splitting criteria in training and the performance evaluation measure in experiments.

Finally, we apply a linear function in each node of MBCT to achieve individual calibration. We observe that traditional binning is typically non-individual, i.e., predicting the same value for all the samples in a bin. This restricts model capacity, making it impossible to improve order accuracy \cite{mcb_bid,bcb_bid}.
By contrast, our linear calibrator function predicts a calibrated value for an individual sample.

To verify the effectiveness and generalization of our method, we conduct comprehensive experiments on three datasets in two application fields. One dataset is constructed by ourselves from real industrial data, and the other two are publicly available benchmark datasets. Experiments show that our method significantly outperforms competing methods in calibration error, and even improves the order accuracy and classification accuracy on certain datasets. Our MBCT is deployed in an online advertising system, achieving significant improvements on business metrics. 

In general, our main contributions are four-fold: 
1) We propose MBCT, a novel calibration method that combines binning and scaling methods. It improves the binning scheme by adopting multiple boosting trees and learning the bias pattern of the predictor with data. It also adopts elements of scaling methods, as we fit a linear function to achieve individual calibration, breaking the bottleneck of traditional calibration methods on order accuracy.
2) We propose a MVCE metric for calibration evaluation. It evaluates the calibration error from multiple perspectives, and is closer to the theoretical value than previous metrics.
3) We construct a dataset of click-through rate prediction with real industrial logs. It fertilizes the research of uncertainty calibration and click-through rate prediction. 
4) We conduct comprehensive experiments to verify the effectiveness of our method and metric. We further conduct online experiments in a real-world application to verify the impact on uncertainty calibration. Results show great commercial value of uncertainty calibration and our proposed method.

\section{Related Work}
With the increasing demand for reliable and transparent machine learning in various applications, a large amount of calibration research work has recently emerged  \cite{verified,distribution_free,seo2019learning,abdar2020review,rahimi2020post}.
Most of the calibration methods serve as a post-processing step, which learns a mapping function to calibrate the output of an off-the-shelf classifier. 
These methods can be mainly divided into three categories: non-parametric, parametric, and hybrid methods.

\textbf{Parametric methods} often assume a parametric data distribution, and use the calibration data to fit the function for training. The Platt scaling \cite{platt1999probabilistic} uses logistic transformation to modify the model output into a calibrated value. This method is further extended to the temperature scaling \cite{equal_width, temperature_scaling} and Dirichlet scaling \cite{kull2019beyond} that can be applied to multi-classification problems. Beta calibration \cite{beta_calibration} focuses on refining the performance on non-logistic predictors. Scaling methods are data efficient, but largely depend on the correctness of the distribution assumption. According to \newcite{distribution_free}, scaling methods cannot achieve distribution-free approximate calibration.

\textbf{Non-parametric methods} do not have distribution assumptions; examples include binning methods \cite{naeini2014binary,bayesian_binning} and isotonic regression \cite{zadrozny2002transforming}. The most popular binning method is histogram binning \cite{uniform_mass}, which groups samples into multiple buckets (bins). Then for each sample, the average prediction results of its bucket is set to be the calibration result. Most of the binning methods (such as histogram binning) is non-strictly mon otonic. Although binning methods have theoretical guarantees in calibration accuracy \cite{distribution_free,verified}, the non-strict monotonicity may hurt applications that concern order accuracy. Isotonic regression \cite{zadrozny2002transforming} also divides the samples into bins and use an isotonic function for calibration with strictly monotonic constraints. 

To take both advantages of scaling and binning methods, \newcite{verified} propose a \textbf{hybrid} method, called scaling-binning, to balance the sample efficiency and the verification guarantees. \newcite{mix_n_match} develop an ensemble method for both parametric and non-parametric calibrations. In our work, we also consider making full use of the parametric and non-parametric methods. Unlike \newcite{verified} and \newcite{mix_n_match}, we adopt the binning framework because of its theoretical guarantee, but  introduce parametric functions to our approach in consideration of order accuracy.

The \textbf{evaluation method} is another key issue in the research of calibration. Most studies use expected calibration error (ECE) and maximum calibration error (MCE) \cite{equal_width,naeini2014binary,bayesian_binning}, but they are shown to be biased and usually underestimated \cite{verified}. 
\newcite{ece_sweep} point out that the bucketing scheme, the number of buckets, and the amount of data will all affect ECE's stability, and claim that
a good metric for calibration should not be so sensitive as ECE. So they propose a strategy to determine the partition method and bin number for obtaining a more stable metric (named $\operatorname{ECE}_{\text{sweep}}$). Experiments show that $\operatorname{ECE}_{\text{sweep}}$ is less biased than existing metrics, such as ECE and MCE. In our paper, We further explore the calibration evaluation methods and propose the multi-view calibration error (MVCE).

\section{Problem Formulation}\label{sec_problem_formulation}
In Sections~\ref{sec_problem_formulation} and~\ref{sec_approach}, we formulate the problem in the binary classification setting for simplicity; the extension to multi-class classification is straightforward. 

Let $\mathcal{X}$ and $\mathcal{Y}=\{0,1\}$ denote the feature and label spaces, respectively. Let $f:\mathcal{X}\to \mathcal{Y}$ be a predictor trained on the training set $D_\text{train}=\{(X_i,Y_i)\}_{i=1}^n$, where $X_i$ and $Y_i$ are drawn i.i.d from data distribution. %$P_{X*Y}$ and $\{1:n\}$ denotes $\{1,2,...,n\}$. 
Note that $Y$ is a random variable even when $X$ is given, and we would like $f$ to predict the expectation $\mathbb E[Y|X]$. The calibration error of $f$ with respect to the ground-truth under $\ell_p$-norm is defined as
\begin{equation}\label{tce}
    \text{TCE}_p(f) = (\mathbb E_X[|\mathbb E[Y|X]-f(X)|^p])^{\frac{1}{p}}
\end{equation}
Typically, $p$ is set to 2~(see \cite{ece_sweep}).
The prediction $f$ is \textit{perfectly calibrated} if the calibration error is zero. In practice, most predictors $f$ (such as logistic regression and deep networks) are not perfectly calibrated and one usually adopts a post-hoc calibrator $h$, which is trained on a calibration set $D=\{(X_i,Y_i)\}_{i=1}^m$. 

According to \newcite{distribution_free}, perfect calibration is impossible in practice, but approximate and asymptotic grouped calibration are possible for finite partitions of $\mathcal{X}$. Here, we restate them as follows for our subsequent analysis:
\begin{definition}[Approximate Grouped Calibration] Suppose $\mathcal{X}$ is divided into $B$ partitions, denoted as $\{\mathcal{X}_b\}_{b=1}^B$. A calibrator $h$ is $(\varepsilon_b,\alpha)$-approximately grouped calibrated for the $b$th group for some $\alpha\in(0,1)$ and $\varepsilon_b\in[0,1]$, if with probability at least $1-\alpha$, 
\begin{equation}\label{app_cal}
    |\mathbb E[Y_b|X_b]- \mathbb E[h(f(X_b))]|\leq \varepsilon_b
\end{equation}
for every $X_b\in\mathcal {X}_b$.
\end{definition}
%$|\mathbb E[Y_b|E(X_b)]- h(f(E(X_b)))|\leq \varepsilon_b(f(E[X_b]))$
\begin{definition}[Asymptotic Grouped Calibration] A calibrator $h$ is asymptotically calibrated  at level $\alpha\in(0,1)$ if, for every $b\in\{1,\cdots, B\}$, $h$ is $(\varepsilon_b,\alpha)$-approximately calibrated for some $\varepsilon_b\in [0,1]$ such that $\varepsilon_b=o_P(1)$, i.e., $\varepsilon_b$ converges in probability to 0 with the size of $D$ increasing.
\end{definition}

\section{Approach}\label{sec_approach}

Our method generally follows the binning framework, but explores a better way of partitioning and further develops individual and non-monotonic calibration to break the bottleneck of traditional calibration methods. Concretely, we adopt Multiple Boosting Trees to learn a feature-aware binning scheme that optimizes a multi-view calibration loss. Our approach is individual but non-monotonic calibration, which is important to order-sensitive applications.
We call our method \textbf{M}ultiple \textbf{B}oosting \textbf{C}alibration \textbf{T}rees (MBCT).

In the following subsections, we first introduce the weakness of traditional binning methods and present feature-aware binning to mitigate this problem. Then, we discuss the necessity of individual and monotonic calibration, and finally, we explain our MBCT method in detail.

\subsection{Feature-Aware Binning}\label{sect_feature_aware}
Previous work~\cite{distribution_free} has proved that binning-based methods can achieve distribution-free approximate grouped calibration and asymptotic grouped calibration. It also provides a theoretical bound for each partition (also referred to as a \textit{bin}):

\begin{theorem}\label{bound_th}
Let $D_b$ denote the set of $D$ samples that fall into the partition $b$. For any $\alpha\in (0,1)$, partition $b$, predictor $f$ and binning-based calibrator $h$, we have with probability at least $1-\alpha$, 
\begin{equation}\label{bound_eq}
    |\mathbb E[Y_b]-\hat{y}_b|\leq \sqrt{\frac{2\hat{V}_b \ln(3B/\alpha)}{c_b}}+\frac{3\ln(3B/\alpha)}{c_b}
\end{equation}
where $\hat{y}_b=\frac{1}{c_b}\sum_{i:X_i\in D_b} Y_i$,  $\hat{V}_b=\frac{1}{c_b}\sum_{i:X_i\in D_b} (Y_i-\hat{y}_b)^2$, $c_b$ is the number of samples in the $b$th bin, and $B$ is the number of bins. For binning methods, we have $\hat{y}_b=\frac{1}{c_b}\sum_{X_i\in D_b} h(f(X_i))$.

\end{theorem}

Note that the guarantee provided by Theorem~\ref{bound_th} is distribution-free. On the contrary, scaling-based methods can only provide theoretical guarantee with distribution assumptions; thus, scaling-based methods are only suitable for specific tasks. 

Although binning-based methods appear to be more general and have some theoretical guarantee, such a guarantee is only for specific partitions determined by the calibration method.
\textbf{In the following, we explain the shortcomings of traditional binning methods in detail and propose feature-aware binning to address the limitations.}

Suppose $h$ is a binning-based calibrator, $h$ will have a division scheme $\operatorname{div}_h$ that divides a dataset into finite subsets (bins). A function $g_b$ is learned to calibrate each bin by minimizing the expected partition calibration error.

\begin{definition}\label{pce_and_epce}
For binning-based methods, a dataset $D$ will be divided into $\mathcal D^{(h)}=\{D_b\}_{b=1}^B$ by the division scheme $\operatorname{div}_h$ of $h$.
For a sample set %$D_b=\{(X_i,Y_i,E(Y_i,|X_i),g_b(f(X_i))\}_{i\in \{1:t\}}$ in $D^{(h)}$, 
$D_b=\{(X_i,Y_i)\}_{i=1}^{c_b}$ in $\mathcal D^{(h)}$, 
the \textbf{partition calibration error} is
\begin{equation}\label{pce}
\operatorname{PCE}(D_b)=|\hat{y}_b^{\operatorname{pred}}-\hat{y}_b|
\end{equation}
We may also define \textbf{expected partition calibration error} (EPCE) as:
\begin{equation}\label{ece_bin}
    \operatorname{EPCE}(D_b)=|\hat{y}_b^{\operatorname{pred}}-\mathbb E[Y_b]|
\end{equation}
where $\hat{y}_b^{\operatorname{pred}}=\frac{1}{c_b}\sum_{X_i\in D_b} g_b(f(X_i))$. 
\end{definition}
Note that the parameters for each bin $b$ may be different, so we use $g_b(f(\cdot))$ to denote the calibration with respect to $b$. We may abbreviate it as $g(f(\cdot))$ if there is no ambiguity. Binning methods ensure that 
$\operatorname{PCE}(D_b)=0$ for every $D_b \in \mathcal D^{(h)}$.

According to Definition~\ref{pce_and_epce} and Theorem~\ref{bound_th}, we have $\text{EPCE} \leq \text{PCE} + |\mathbb E[Y_b]-\hat{y}_b| \leq \sqrt{\frac{2\hat{V}_b \ln(3B/\alpha)}{c_b}}+\frac{3\ln(3B/\alpha)}{c_b} $. If the variance of $D_b$ is finite and its size $c_b$ is large enough, then $\text{EPCE}(D_b)\approx 0$. 
This ensures that the binning-based methods can still optimize the EPCE of the division even if the ground-truth cannot be obtained.

\begin{definition}\label{def_under_degree}
For a bin $D_b$, the \textbf{partition underestimated degree} (PUD) with respect to $f$ and $h$ is:
\begin{equation}
    \operatorname{PUD}(D_b)=\hat{y}_b^{\operatorname{pred}}/{\hat{y}_b}
\end{equation}
Similarly, for one sample $X_i$, we define the sample underestimated degree (SUD) as:
\begin{equation}
    \operatorname{SUD}(X_i)={g(f(X_i))/{\mathbb E[Y_i|X_i]}}
\end{equation}
\end{definition}
Apparently, the bin $D_b$ is underestimated when $\text{PUD}(D_b)<1$, or overestimated when $\text{PUD}(D_b)>1$. For SUD, it is not possible to calculate its value in practice, it will facilitate our theoretical analysis.

We say a bin $D_b$ is ($\beta$, h)-\textbf{well-calibrated} when $\operatorname{EPCE}(D_b)\leq \beta$.
If $\beta$ is small enough in real scenarios, we say $D_b$ is a \textbf{well calibrated partition}.

For a bin $D_b$, we may further split it by a division scheme $\operatorname{div}$ into $k$ equally sized subsets $\{D_{b_i}\}_{i=1}^k$. For simplicity, we define the $(k,\operatorname{div})$-\textbf{balanced finer-grained partition calibration error} (BFGPCE) of $D_b$ as:
\begin{equation}\label{fine_grain_loss}
    \operatorname{BFGPCE}_{k,\operatorname{div}}(D_b)=\frac{1}{k}\sum_{i=1}^k|\hat{y}_{b_i}^{\operatorname{pred}}-\hat{y}_{b_i}|
\end{equation}
Further, \textbf{$(k,\operatorname{div})$-balanced finer-grained expected partition calibration error} (BFGEPCE) is defined as
\begin{equation}\label{expected_fine_grain_loss}
    \operatorname{BFGEPCE}_{k,\operatorname{div}}(D_b)=\frac{1}{k}\sum_{i=1}^k|\hat{y}_{b_i}^{\operatorname{pred}}-\mathbb E[y_{b_i}]|
\end{equation}

For a certain well-calibrated subset $D_b$, $\operatorname{BFGPCE}_{k,\operatorname{div}}(x_b)$ and $\operatorname{BFGEPCE}_{k,\operatorname{div}}(x_b)$ may still be large if a different division is used, in which case the calibration performance may not be satisfactory.  
Then, a natural question follows: Is it possible to achieve near-perfect 
calibration based on the binning method? %, or what is the best possible state of the binning method? 
Perfect calibration under binning methods can be restated as: For each bin $D_i\in \mathcal D^{(h)}$, (1) $D_b$ is well calibrated under $h$, and (2) each sample $(X_i,Y_i)\in D_b$ satisfies $\mathbb E[Y_i|X_i]=g_b(f(X_i))$. Here, we call $g_b$ the bias pattern of the bin $D_i$. Notice that condition (1) can be achieved by all the binning-based methods with sufficient samples. Condition (2) is a sufficient condition for (1). Thus, if we can put the samples that have the same bias pattern into the same bin, we will achieve a near-perfect calibration.

Traditional binning methods aim to find a constant bias pattern and usually assume that the samples' groundtruth probabilities are approximately the same if the predicted probabilities by $f$ are close; hence, the samples are sorted by the predicted probabilities and divided into partitions in an  uniform-mass~\cite{uniform_mass} or equal-width~\cite{equal_width} manner. 
However, the assumption of traditional binning-based methods is too strong. In fact, the data may deviate from this assumption; especially, the difference is large when the estimation accuracy is low. \newcite{verified} also prove that there exist better partition schemes than uniform-mass and equal-width methods with sorted samples, although they do not provide a method to find them.

In this paper, we aim to find a better partition scheme for binning-based calibration. We notice that, if $h$ can achieve approximately perfect calibration, we will have the following property: for any subset $D_{sub}$ (with sufficient samples) of $D$, 
we get $\operatorname{EPCE}(D_{\text{sub}})\approx 0$. In other words, the more partitions with $\text{EPCE}\approx 0$, the closer to perfect calibration $h$ would be. Thus, we propose the multi-view calibration error for evaluating the calibration with respect to a set of partitions $\{\operatorname{div}_i\}_{i=1}^r$:
\begin{equation}\label{mvce}
    \operatorname{MVCE}_{f,h,\{\operatorname{div}_i\}_{i=1}^r}(D_b)=\Big(\frac{1}{r}\sum_{i=1}^r\Big(\frac{1}{t_i}\sum_{j=1}^{t_i} \operatorname{PCE}(D_{i,j})\Big)^p\Big)^{\frac{1}{p}}
\end{equation}
where $t_i$ is the number of bins under $\operatorname{div}_i$ and $D_{i,j}\in \operatorname{div}_i(D)$. Intuitively, MVCE is suitable for measuring how close the calibration is to the optimum. 
Another intuition is the bias between $f(X)$ and $\mathbb E[Y]$ would have different widespread patterns related to $X$ that can be captured by machine learning methods. Therefore, \textbf{we propose a feature-aware binning framework to learn which samples belong to a  bias pattern}. It employs a model $M$ which takes the features $X$ as input and optimizes the MVCE. Optimizing MVCE can force the model to put the samples with a similar bias pattern into the same bin.

Specifically, we employ a linear function $g(f(\cdot),b)=k_bf(\cdot)$ as the bias pattern to achieve individual calibration~(Section~\ref{subsec_indi}). In other words, we would like to find samples that have the same SUD. Details of our methods are presented in Section~\ref{subsec_mbct}.

\subsection{Individual and Monotonic Calibration}
\label{subsec_indi}
\begin{figure*}[h]
  \centering
  \includegraphics[width=\linewidth]{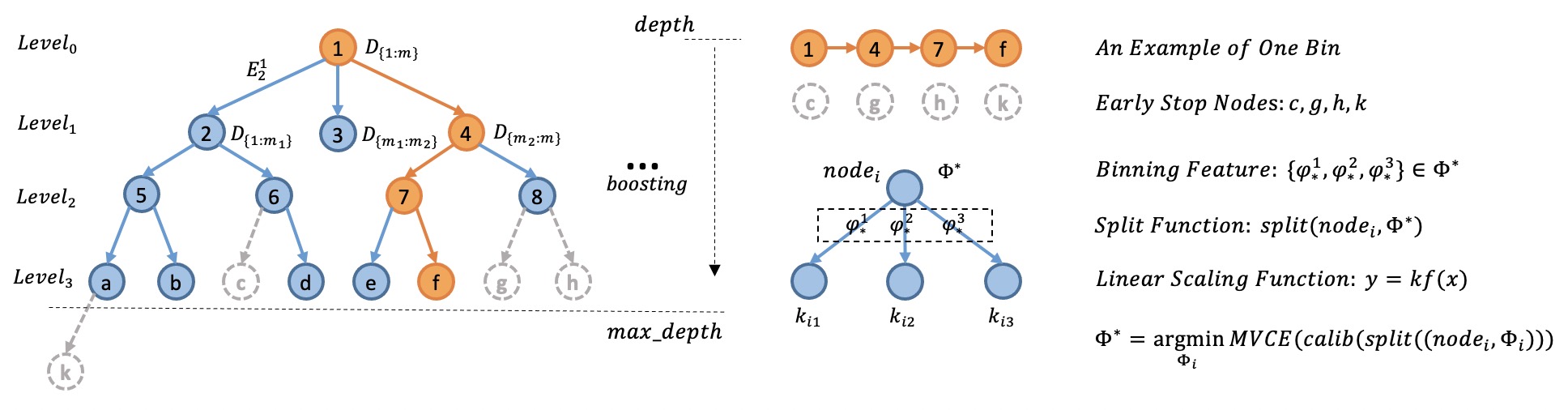}
  \caption{An illustration of our Multiple Boosting Calibration Trees (MBCT). The left-hand side shows an example of a binning tree. MBCT may have multiple boosting trees to minimize our MVCE loss. The right-hand side shows several key elements and concepts in our MBCT. Consider a node $i$ with corresponding data $D_i$. It selects a feature $\Phi_*$ by MVCE, and splits the node by the feature values $\varphi_*^1, \varphi_*^2,\varphi_*^3$. For each child node, a linear function $g_j(x)=k_jx$ is fitted for calibration.}\label{fig_mbct}
\end{figure*}

Besides the true probability, 
order accuracy, often measured by the area under the curve (AUC), is also important to various applications, such as computation advertising~\cite{tdm,dien} and web search~\cite{websearch_borisov2018calibration}.
Traditional binning-based methods are non-strictly monotonic because they set the same calibrated for samples in the same bin, which is also considered as non-individual. In this part, we first give the definitions of individual and monotonic calibration and then discuss their necessity in uncertainty calibration.

\textbf{Individual} calibration means that we predict a calibrated value for an individual sample. By contrast, a \textit{non-individual} calibrator predicts a value for a set of pooled samples.

\textbf{Monotonic} calibration means that the order is preserved. Concretely, for a dataset $D=\{(X_i,Y_i)\}_{i=1}^m$, we call $h$ a \textbf{non-strictly monotonic} calibrator if
for every predictor $f$ and every two samples $(X_i,Y_i), (X_j,Y_j)\in D$, we have:
\begin{equation}\label{mono_condition}
    (f(X_i)-f(X_j))\cdot (h(f(X_i))-h(f(X_j)))\ge 0
\end{equation}

If the equality holds only when $f(X_i)=f(X_j)$, then $h$ is a \textbf{strictly monotonic} calibrator. 

If there exists two samples violating Equation~(\ref{mono_condition}), we say $h$ is a \textbf{non-monotonic} calibrator.

In practice, non-individual calibration is harmful to order-sensi-tive applications, such as computation advertising. For example, we would like to select samples with top-3 probabilities from a sample set; non-individual calibrators may give 10 samples in a bin with the same predicted probability. Thus, individual calibration is crucial to order-sensitive uncertainty calibration, and our MBCT  makes samples distinguishable by learning a linear function $g$ for a bin.

On the other hand, traditional methods often apply non-strictly or strictly monotonic calibrators, assuming that the base classifier $f$ can predict the correct order. However, this assumption appears too strong, and restricts the power of the calibrator. In our work, we lift such constraint and our MBCT may be non-monotonic.

\subsection{Multiple Boosting Calibration Trees}\label{subsec_mbct}

As mentioned, our paper proposes a feature-aware binning framework, which adopts a machine learning model $M$ to optimize the MVCE loss; our calibrator should be individual but non-monotonic. 

Concretely, we adopt multiple boosting trees as the model $M$ mainly for the
following considerations: (1) Tree-based models are more friendly to interpretation, and it is natural to assign a bin to each node in the tree. (2) It is easy to optimize custimized losses (such as MVCE) with boosting trees. (3) Boosting methods are shown to be effective in various applications~\cite{adaboost,chen2016xgboost}.

Figure~\ref{fig_mbct} shows an overview of our approach, Multiple Boosting Calibration Trees (MBCT). It takes discretized features as input. Let $\mathscr F=\{\Phi_i\}_{i=1}^d$ be the feature space, each feature $\Phi_i$ taking values $\varphi_i^1,\cdots, \varphi_i^a$.  We adopt the random shuffle uniform-mass partition scheme as division of MVCE, i.e., we first randomly shuffle the samples (instead of sorting by the predictions), and then perform a uniform-mass division. This enables us to obtain multiple uniform-mass divisions.

There are three key operations during the growth of a binning tree: (1) For each leaf node, select the best feature to split the node. For a node with its data $D'$, the best feature is obtained by
\begin{equation}\label{select_feat}
    \Phi_*=\operatorname{argmin}\limits_{\Phi_i\in\mathscr F} \operatorname{MVCE}(\operatorname{calib}(\operatorname{split}(D',\Phi_i))))
\end{equation}
where $\operatorname{split}(\cdot)$ is an operation that splits a node in the calibration tree by the values of $\Phi_i$, and $\operatorname{calib}(\cdot)$ calculates the calibration parameters (in our model, it is a linear function). (2) Split the leaf node by its $\Phi^*$ and calibrate its children, if the early stop conditions are not met and $\Phi^*$ can split the data of the node at least into 2 subsets. (3) If there is no leaf node can be extended, finish the growth of the binning tree. 

The stopping conditions for growing one tree include: (1) The node reaches the maximum depth set as a hyper-parameter. (2) The samples in the node is less than $\beta$, which is heuristically estimated by Equation~(\ref{bound_eq}). Concretely, we first calculate $\hat{V}_D$ and $\hat{y}_D$ by Equation~(\ref{bound_eq}) on the whole dataset $D$. Then, we assume MBCT is close to a uniform-mass binning which satisfies $B=\frac{|D|}{c}$, where $c$ is the sample number in each bin. Finally, given a permissible mean absolute percentage error limit $e$ and confidence $\alpha$, we obtain the maximum $c$ that satisfies:
\begin{equation}\label{mbct_app_c}
    \hat{y}_D\leq \frac{1}{e}(\sqrt{\frac{2\hat{V}_D \ln(3|D|/c\alpha)}{c}}+\frac{3\ln(3|D|/c\alpha)}{c})
\end{equation}
We set $\beta$ as $c$, and we simply set the bin size of the loss as $\frac{c}{2}$ because we need at least 2 bins to calculate the calibration loss. $\alpha$ and $e$ are hyper-parameters for our method. (3) The local loss rises after splitting and calibrating the node. Here, the local loss refers to the loss on the subset corresponding to the node to be divided and calibrated. We do not adopt global loss because it will greatly increase the complexity for MBCT. 

MBCT allows to generate multiple binning trees. We decide whether to generate a new binning tree according to the change of global loss. 
A path from the root node to the leaf node represents a specific bin. Multiple binning trees are actually doing the ensemble of different binning schemes for further optimizing the calibration loss. With these operations, MBCT optimizes calibration loss in a greedy way. Although greedy optimization may not be optimal, it is considered as a very effective optimization algorithm in various models, including boosting trees.

\section{Experiments}

\subsection{Experimental Setup}
We first conducted simulation experiments to verify the rationale for using MVCE to evaluate calibration errors, which is also applied to our online advertising system for calibrating the predicted Click-Through Rate (pCTR). Then, we conducted both offline and online experiments to verify the effectiveness of our method and the impact on the real-world application. 

For the offline experiment, we dumped the impression and click logs from our commercially deployed online display advertising system, and we call our industrial dataset CACTRDC\footnote{ https://github.com/huangsg1/Tree-Based-Feature-Aware-Binning-for-Individual-Uncertainty-Calibration
}, which is the acronym for Computation Advertising Click-Through Prediction Dataset for Calibration.

In addition, we tested our method on two public datasets, Porto Seguro\footnote{https://www.kaggle.com/c/porto-seguro-safe-driver-prediction} and Avazu\footnote{https://www.kaggle.com/c/avazu-ctr-prediction}, showing the generality of our method.
Table~\ref{stat} presents the statistics of these datasets.

\begin{table}[!t]
\centering	
\caption{Dataset statistics. ``Predictor Train'' refers to the training set of the base predictor $f$.  ``Calibration Train'' and ``Calibration Test'' are the training and test sets for the calibrator $h$, respectively. \label{stat}}
\vspace{-2mm}
    \resizebox{0.95\linewidth}{!}{
 		\begin{tabular}{c|c|c|c}
 			    \thickhline
 		     Dataset & Predictor Train & Calibration Train	& Calibration Test \\     \thickhline
 		    CACTRDC & 48M & 11M & 1M\\ \hline
 		    Porto Seguro & 357K & 208K & 30K \\ \hline
 		    Avazu   & 24M & 12M & 4M \\
 		         \thickhline
 		\end{tabular}}

\end{table}

Regarding the base predictor, we used a deep neural network (DNN) for CACTRDC and our online A/B testing, as already deployed in the system. For the Porto Seguro and Avazu datasets, we use the Gradient Boosting Decision Tree (GBDT), because we would like to experiment with non-DNN models. In principle, our calibration is agnostic to the base predictor; this further demonstrates the generality of our calibration approach.

In our experiments, we conducted feature engineering and tuned hyperparameters on the CACTRDC dataset, which is related to our deployed commercial system. For the Porto Seguro and Avazu benchmark datasets, we mostly adopted the same settings, as the main focus of our approach is not hyperparameter tuning; this, again, shows the generality of our MBCT. 
For discrete features, we directly take their values when splitting a node by the feature; continuous features are discretized by quantiles. We also convert the outputs of the predictor into discrete features for MBCT. More feature engineering details are shown in appendix~\ref{app:feature_engineering}.

For hyperparameters, we set $\alpha$ to 0.05 and $e$ to 0.1 (Section~\ref{subsec_mbct}), so the bin sizes were 800, 2000, and 1000 for Porto Seguro, Avazu, and CACTRDC, respectively. The max tree depth was 5 and the max tree number was 8. 

For evaluation of the calibration methods, we adopt MVCE for calibration accuracy and area under the curve (AUC) for order accuracy. We set the $r$ of MVCE to 100.

\subsection{Competing Methods}
We compare our MBCT model with the following state-of-the-art methods in previous studies.

\begin{itemize}[leftmargin=*]
\item \textbf{Platt Scaling.} \newcite{platt1999probabilistic} propose a scaling method that uses logistic function to transform the model predictions into calibrated probabilities.
\item \textbf{Beta Calibration.} \newcite{beta_calibration} refine the assumption of isotonic calibration, which is more suitable for non-logistics predictors, such as Na\"ive Bayes and AdaBoost.
\item \textbf{Histogram Binning.} \newcite{uniform_mass} propose a method that divides the samples into bins and calibrates each sample by the average probability of its bin as the calibrated value.
\item \textbf{Isotonic Regression.} \newcite{zadrozny2002transforming} propose a method that learns an isotonic function for calibration with the strictly monotonic constraint.
\item \textbf{Scaling-Binning.} \newcite{verified} propose a scaling-binning calibrator that combines the Platt scaling and Histogram binning. The approach first fits a parametric function to reduce variance and then puts the function values into bins for calibration. Note that it is a non-individual and non-strictly monotonic calibrator, which is different from our method.
\end{itemize}

\subsection{Analysis of the Metrics for Calibration Error}\label{sim_mvce}
As stated in Section~\ref{sect_feature_aware}, MVCE is intuitively an appropriate metric for evaluating calibration error. Also, we notice that previous work often uses ECE or $\text{ECE}_{\text{sweep}}$ to estimate the calibration error. ECE with $n$ bins in $\ell_p$-norm can be formulated as:
\begin{equation}\label{ece_n}
    \operatorname{ECE}_n=\big(\frac{1}{N}\sum_{i=1}^N \operatorname{PCE}(D_i)^p\big)^{\frac{1}{p}}
\end{equation}
where $D_i$ represents the samples in the bin $i$, and $n$ is a hyper-parameter referring to the number of the bins for evaluation.
Note that in ECE samples are sorted by calibration results, based on which uniform-mass binning is performed. $\operatorname{ECE}_{\text{sweep}}$ is proposed by \newcite{ece_sweep} and has a strategy to determine the partition method and bin number, which is shown to be less biased than ECE. 

Intuitively, our MVCE evaluates the calibration error from a more comprehensive perspective; ECE and $\text{ECE}_{\text{sweep}}$ are special cases in MVCE. 
In this part, we will evaluate these metrics with a simulation experiment.  
In fact, the metric for calibration ultimately concerns the theoretical calibration error (TCE), which unfortunately cannot be computed on real-world data, because the true probability of a sample in unknown. Using simulated data with certain assumptions allows us to compute TCE and evaluates a calibration metric.

Therefore, we design a simulation experiment, where we assume $f(X)$ follows a certain known distribution and bias pattern, so that the TCE can be computed. 
Concretely, we assume $h(X)\sim \operatorname{Beta}(0.2,0.7)$ 
 and $\mathbb E[Y|h(X)=c]=c^2$, where we hide the role of $X$ as it is irrelevant to our simulation.
 Then, we can compute the TCE according to Equation~(\ref{tce}) with the $\ell_2$-norm. The TCE in our simulation experiment is 0.0868.

We evaluate a metric $\mu$ (e.g., ECE, ECE$_\text{sweep}$, and MVCE; also with $\ell_2$-norm) by comparing it with TCE. The expected difference for a sample $X$ is formulated as:
\begin{equation}\label{e_bias}
    E_{bias}= |\mathbb E[\mu(X,h)]-\operatorname{TCE}(X,h)|
\end{equation}

We used the Monte Carlo method to estimate $E_{bias}$, where we conducted $m=200$ experiments, each with random $n$ samples. The empirically estimated difference is given by
\begin{equation}\label{e_bias_hat}
    \hat{E}_{bias}(n)=\frac{1}{m} \sum_{i=1}^{m} |\mu{(X_i^{(n)},h)}-\operatorname{TCE}{(X,h)}|
\end{equation}
where $X_i^{(n)}$ means a set of $n$ samples.

\begin{figure}[!t]
  \centering
  \includegraphics[width=1.0\linewidth]{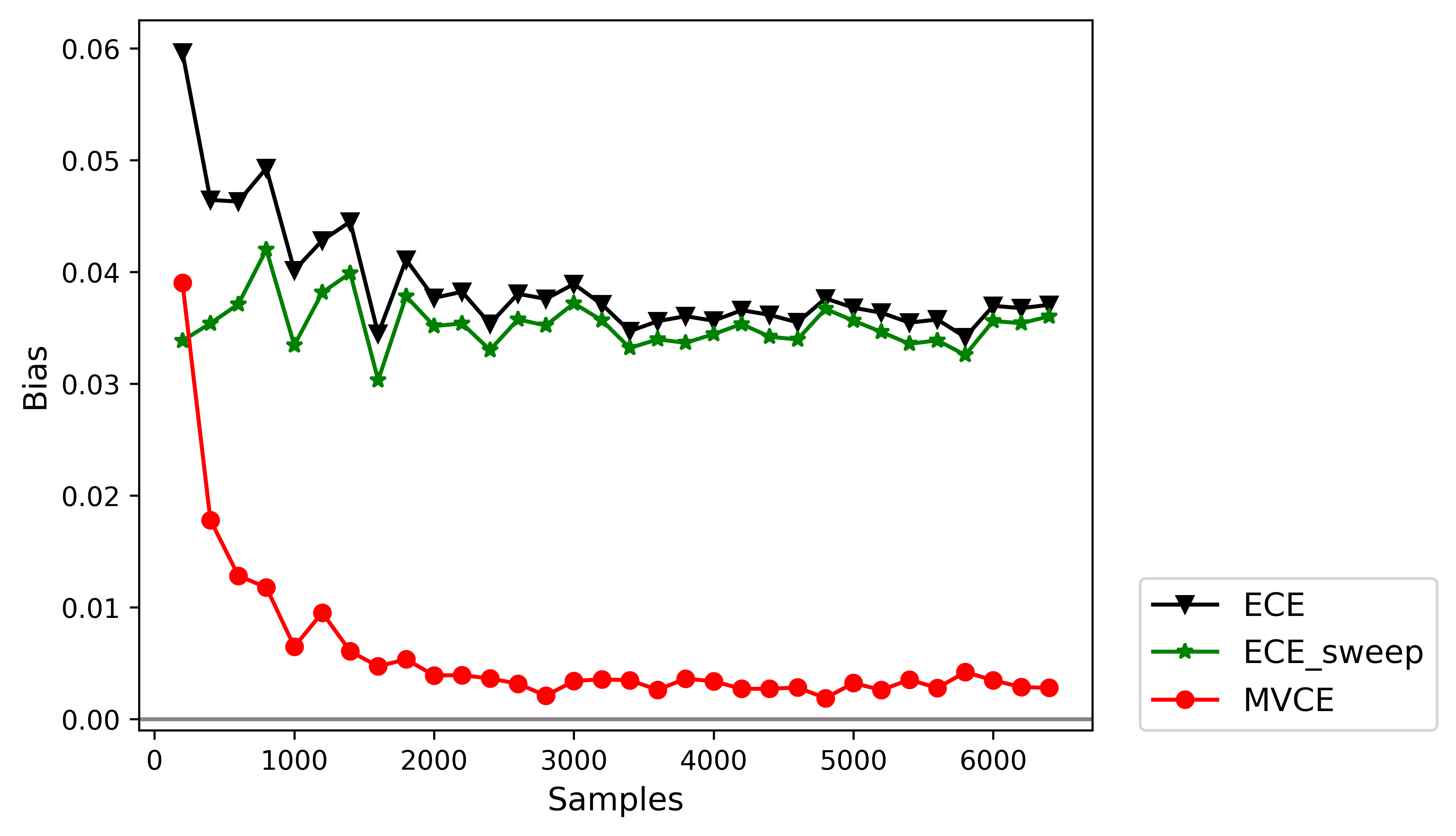}\vspace{-2mm}
  \caption{Main simulation results of ECE, $\text{ECE}_{\text{sweep}}$ and MVCE. The bin numbers of ECE and MVCE are set as 32 in this experiment. The bin number of $\text{ECE}_{\text{sweep}}$ is fixed and determined by itself.}\label{fig_mvce_sim_curve}
  \vspace{-2mm}
\end{figure}

Figure~\ref{fig_mvce_sim_curve} shows the estimated difference of ECE, $\text{ECE}_{\text{sweep}}$, and MVCE under various sample sizes $n$. We see that $\text{ECE}_{\text{sweep}}$ is better than ECE, which is consistent with the experimental conclusion of \newcite{ece_sweep}. MVCE achieves the best result, indicating that MVCE is a better metric for calibration error than ECE and $\text{ECE}_{\text{sweep}}$.

\begin{figure}[!t]
  \centering
  \includegraphics[width=1.0\linewidth]{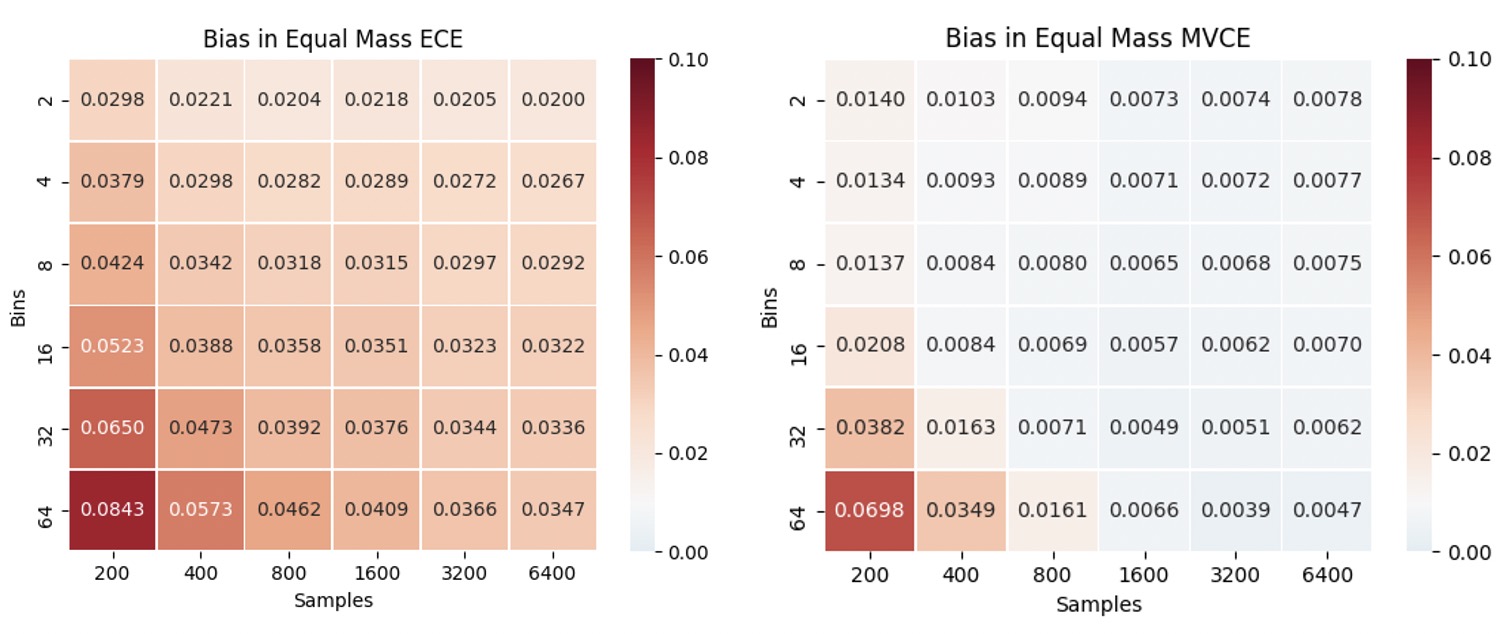}\vspace{-2mm}
  \caption{$\hat{E}_{bias}$ of MVCE and ECE under different numbers of bins and samples.}\label{fig_mvce_sim_heat_map}
\end{figure}
%\vspace{-1mm}

We further study the impact of different combinations of bin and sample numbers on ECE and MVCE in terms of $\hat{E}_{bias}$ in Figure~\ref{fig_mvce_sim_heat_map}. We see that (1) MVCE is consistently better than ECE in different cases; that (2) more data lead to less bias for ECE under a certain bin number, but the larger the amount of data, the smaller the effect is; and that (3) the bin number should be moderate for MVCE with respect to the sample number, and neither too large nor too small is desired. We provide more supplement results with different settings in appendix~\ref{app:more_simulation}.

\subsection{Main Results}\label{main_result}
 Table~\ref{exp_main} presents the main results of our approach on three datasets. Numbers in bold indicate the best performance. A higher score of AUC indicates a better model; for MVCE, a lower score is better. 

We see that MBCT consistently outperforms the original base predictor and all the baseline calibration methods in terms of both MVCE and AUC metrics, which indicates that our method can bring improvement on both calibration and order accuracy, and is generalizable to different datasets. The AUC of other methods fluctuates after calibration, which is because these methods may turn different estimates into exactly the same.

\begin{table}[t]
    \caption{Results for uncertainty calibration on the benchmark datasets. 
    \label{exp_main}}
 	\centering	
 	\resizebox{1.0\linewidth}{!}{
 		\begin{tabular}{c|c|c|c|c|c|c}
 			\thickhline
 \multirow{2}{*}{Method} & \multicolumn{2}{c|}{CACTRDC} & \multicolumn{2}{c|}{Porto Seguro} & \multicolumn{2}{c}{Avazu} \\ 
 	\cline{2-7}
 & MVCE $\downarrow$ & AUC $\uparrow$ &  MVCE $\downarrow$  & AUC $\uparrow$  &  MVCE $\downarrow$ & AUC $\uparrow$ \\ \thickhline
 	Original Predictions & 0.00394 & 0.77902 & 0.00619 & 0.62869 & 0.00976 & 0.71880 \\ \hline
    Platt Scaling & 0.00374 & 0.77902 & 0.00604 & 0.62869 & 0.00792 & 0.71880
    \\ 
    Beta Calibration & 0.00371 & 0.77902 & 0.00601 & 0.62869 & 0.00789 & 0.71880\\ \hline
    Histogram Bining & 0.00372 & 0.77895 & 0.00597 & 0.62998 & 0.00787 & 0.72381 \\ %\hline
    Isotonic Regression & 0.00371 & 0.77915 & 0.00598 & 0.62936 & 0.00787 & 0.72030 \\ \hline
    Scaling-Binning & 0.00373 & 0.77892 & 0.00605 & 0.62880 & 0.00792 & 0.71871 \\ \hline
    \textbf{Our full MBCT} & \textbf{0.00368} & \textbf{0.78693} &  \textbf{0.00586} & \textbf{0.63097} & \textbf{0.00780} & \textbf{0.74177} \\ 
    \textbf{Our model w/o boosting}  & 0.00374 & 0.78373 & 0.00595 & 0.62999  & 0.00784 & 0.73797 \\
    \thickhline
    %\bottomrule
 	\end{tabular}
}
\end{table}

Moreover, we conduct an ablation study of the boosting strategy in the last two rows of Table~\ref{exp_main}. As seen,  boosting consistently brings in improvement of AUC and MVCE on CACTRDC, Porto Seguro, and Avazu. Even without boosting, our method achieves the best AUC on all the benchmarks and achieves the best MVCE on Porto Seguro and Avazu. 
We notice that boosting brings more improvement of AUC on large datasets; this is probably becasue larger datasets can benefit more from the enlarged model capacity brought by boosting.

\subsection{In-Depth Analysis}\label{in_depth_ana_main}

In this part, we conduct in-depth analysis for our model. Due to the limit of time and space, we take the CACTRDC dataset as the testbed, unless otherwise stated.

\begin{figure}[t]
  \centering
  \includegraphics[width=1.0\linewidth]{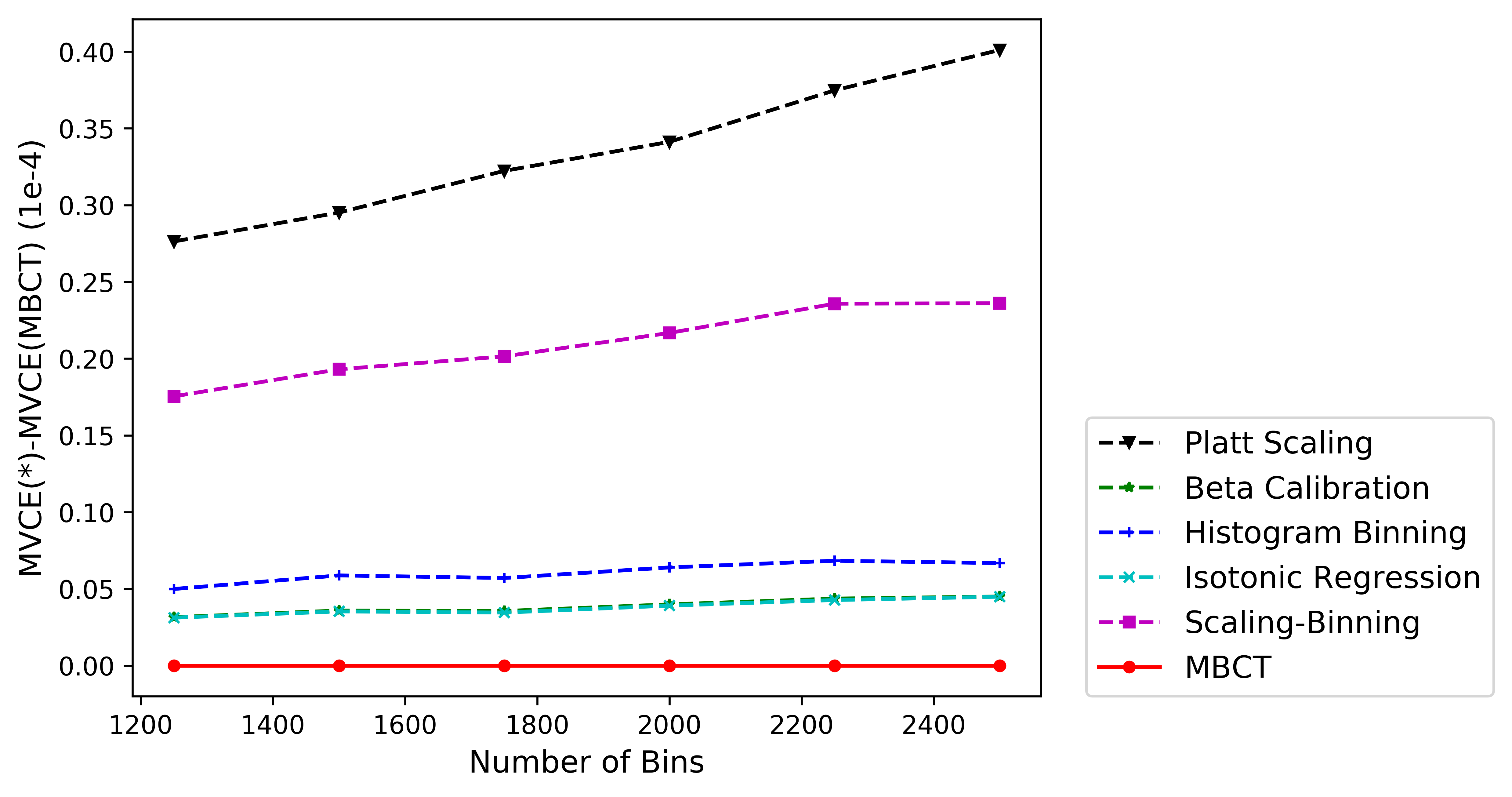}
  \vspace{-.2cm}\caption{The difference of MVCE scores in comparison  with our MBCT. Natually, our approach yields a value of 0. All competing models have a positive value, showing that our approach is the best. }
   \vspace{-.2cm}
  \label{main_result_fig}
\end{figure}

In Table~\ref{exp_main}, we simply set the bin size of MVCE and ECE to the minimum bin size ($\beta$ in Section~\ref{subsec_mbct}) during training. The MVCE score under different bin sizes is also a curious question, and we show the trend in  Figure~\ref{main_result_fig}. We see that MBCT consistently outperforms the baselines on MVCE under all bin numbers, which further illustrates the effectiveness of our method.

\begin{figure}[t]
  \centering
  \includegraphics[width=1.0\linewidth]{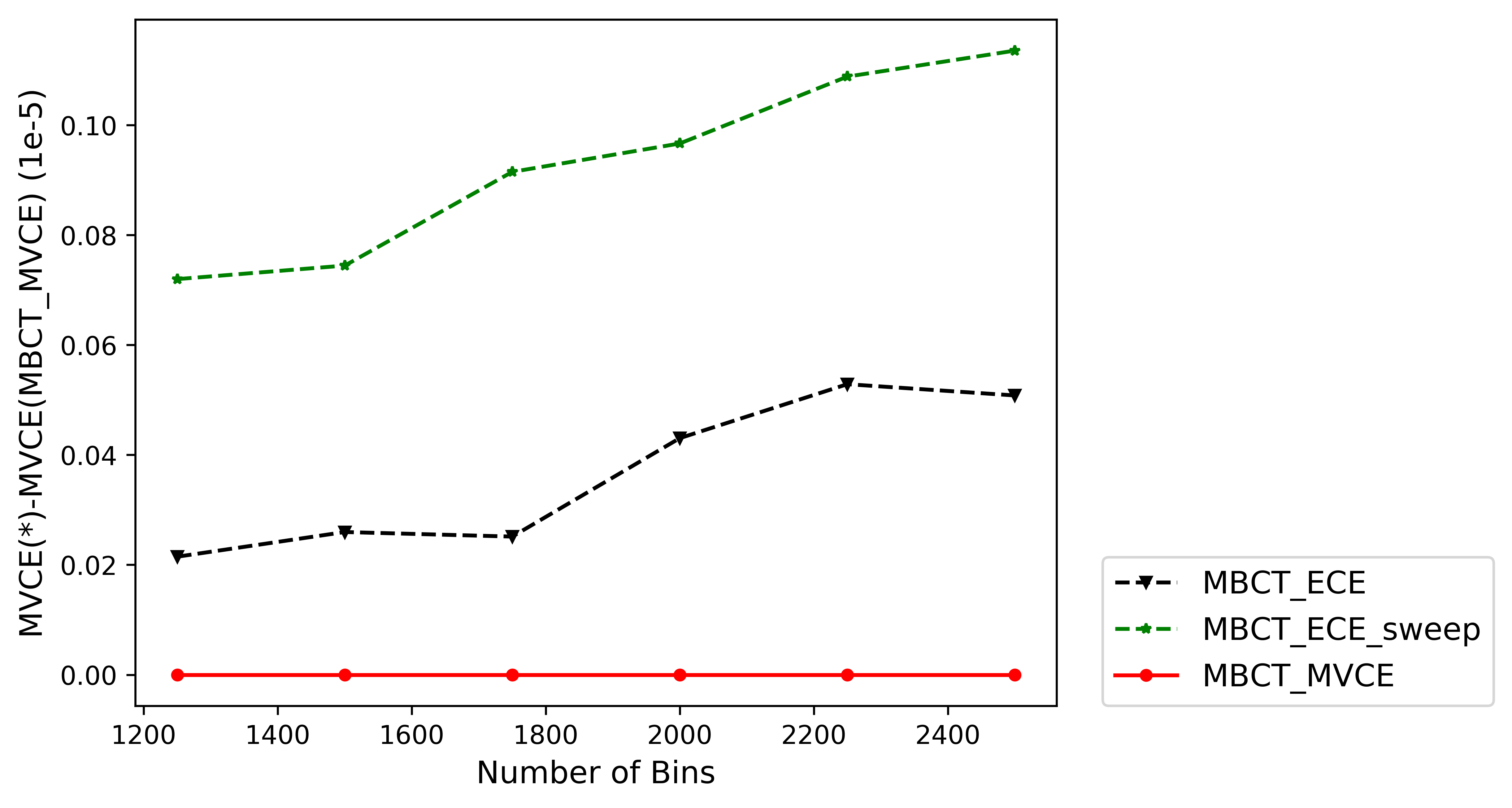}
  \vspace{-.2cm}
  \caption{Ablation study of the training loss (in comparison with MVCE training).}
  \label{mbct_loss_ablation}
\end{figure}

Then, we analyze the effect of training losses. We compare our MVCE loss with traditional ECE and ${\text{ECE}_\text{sweep}}$ losses in Figure~\ref{mbct_loss_ablation}.
We see that using MVCE as the training loss consistently outperforms ECE and $\text{ECE}_{\text{sweep}}$ on the three datasets, which shows that it is beneficial to consider the average PCEs from different perspectives during training. 

\begin{figure}[h]
  \centering
  \includegraphics[width=1.0\linewidth]{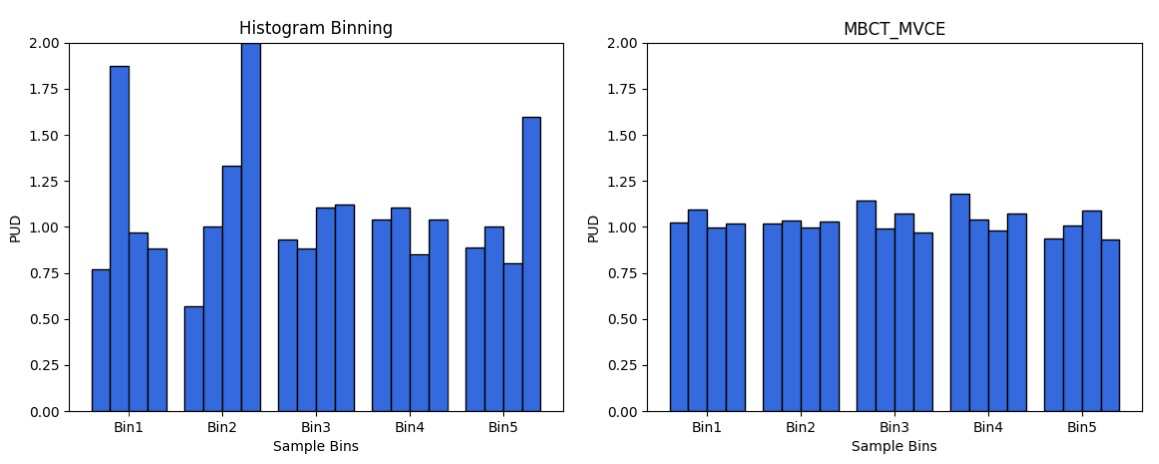}
  \caption{PUD of finer-grained partitions in MBCT and Histogram Binning.}
  \label{pud_visualization}
\end{figure}

We further conduct an analysis to show whether MBCT puts samples with similar bias patterns for $f$ into the same bin. We randomly select 5 bins of MBCT and Histogram methods, and divide each bin into 4 sub-groups. Each bin of MBCT is corresponding to a leaf node. Figure~\ref{pud_visualization} shows the PUD of these sub-groups. It is obvious that the sub-groups of MBCT is closer to 1, which implies that MBCT learns better relationships between the inputs and the bias pattern of $f$ than naive binning methods.

\begin{figure}[h]
  \centering
  \includegraphics[width=1.0\linewidth]{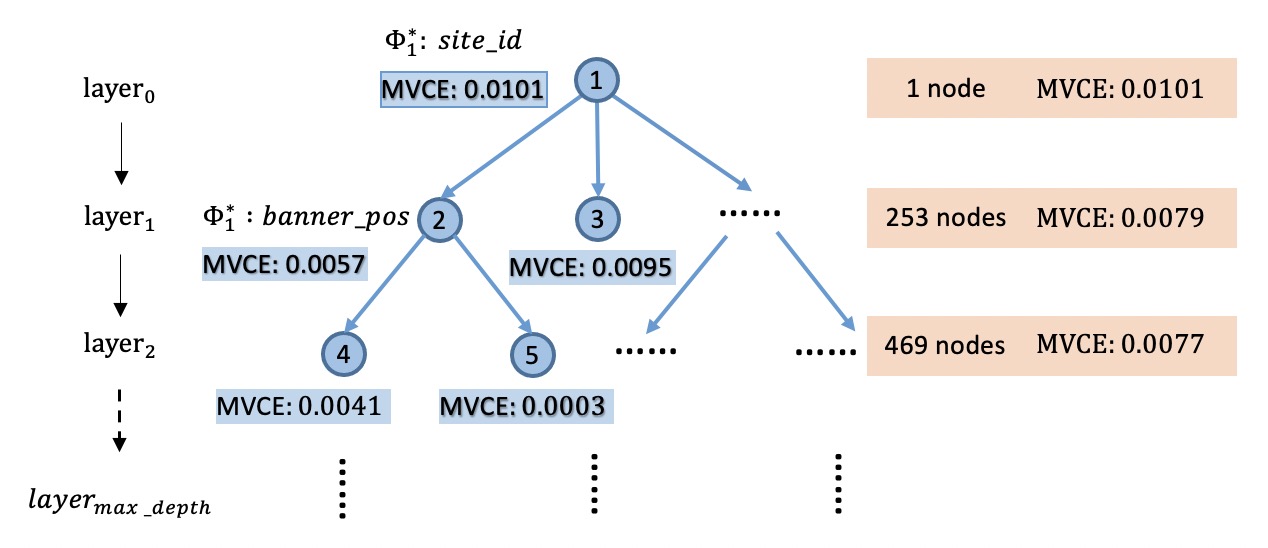}
  \caption{Visualization of the greedy optimization in MBCT. Local MVCE of one node is in blue, and the Global MVCE of the whole dataset is in orange. }
  \label{fig_vis_tree}
\end{figure}

To better understand our MBCT model, we present a case study in Figure~\ref{fig_vis_tree}, showing the growth of the calibration tree. It shows a few nodes in the top 3 layers of the first tree of MBCT. As seen, our approach splits a node by optimizing the local MVCE, and this greedy way can effectively reduce the global loss.

\subsection{The Impact of Calibration on Real-world Applications}\label{sec_online_exp}
To the best of our knowledge, previous work has not applied their calibration methods to real-world applications. It is important to test whether the calibration can improve performance in the real world. We deploy MBCT to calibrate the click-through rate for an online advertising system and conduct an online A/B test for 15 days.
We use click-through rate (CTR) and the effective cost per mille (eCPM) to evaluate the models, which are the key performance metrics:
\begin{equation}
    \text{CTR}=\frac{\text{\#\space~ of\space~ clicks}}{\text{\#\space~ of\space~ impressions}},~ \text{eCPM}=\frac{\text{Ad~revenue}}{\text{\#\space ~of\space~ impressions}}
\end{equation}
We build three buckets on the online traffic platform to conduct fair comparison. We reserve bucket A with no calibration. Isotonic regression and MCBT are applied to buckets B and C, respectively.
Each bucket has 3\% of our online traffic, which is large enough considering the overall page view requests.
Table~\ref{online_result} lists the promotion of the two main online metrics. 5.7\% growth on AUC exhibits that MBCT could improve the order accuracy of the online predictor. As for CTR, it has a 22.1\% improvement, indicating that MBCT can bring more clicks because order accuracy is highly related to the allocation efficiency. Regarding eCPM, a 8.4\% lift demonstrates more revenue for the platform. These improvements are significant for an online advertising system, demonstrating great business value. We provide more details and discuss the relationship between calibration and revenue in Appendix~\ref{add_online_deploy_impl} and ~\ref{app:discussion}.

\begin{table}[H]
\caption{Online results of 15 days in comparison with no calibration and Isotonic Regression.\label{online_result}}
 	\centering	
    \resizebox{0.9\linewidth}{!}{
 		\begin{tabular}{c|c|c|c}
 			    \thickhline
 		    Metric & No Calibration	& Isotonic Regression & MBCT \\     \thickhline
 		    AUC & 0.0\% & +0.3\% & +5.7\% \\ \hline
 		    CTR & 0.0\% & +8.8\% & +22.1\% \\ \hline
 		    eCPM   & 0.0\% & +4.5\% & +8.4\% \\     \thickhline
 		\end{tabular}}
\end{table}

\section{Conclusion}
In this paper, we propose a novel calibration method, called MBCT, which extends the binning framework in several ways. We adopt multiple boosting trees in the feature-aware binning framework, and propose the multi-view calibration error (MVCE) to learn a better binning scheme. We also learn a linear function to calibrate the samples in a bin. Thus, our approach is individual but non-monotonic, which is important to order-sensitive applications.

We conduct experiments on both public datasets and an industrial dataset, covering different application fields. Experimental results show that MBCT significantly outperforms the competing methods in terms of calibration error and order accuracy; profoundly, it outperforms the original prediction in order accuracy, which cannot be achieved by monotonic calibrators. We also conduct simulation experiments to verify the effectiveness of our proposed MVCE metric.  Our approach is deployed in an online advertising system. Results show high performance and great commercial value of our approach in real-world applications. 

\section*{Acknowledgments}
Lili Mou is supported in part by Natural Sciences and Engineering Research Council of Canada
(NSERC) under Grant No.~RGPIN2020-04465, the Amii Fellow
Program, the Canada CIFAR AI Chair Program, and
Compute Canada (www.computecanada.ca).

%%
%% The next two lines define the bibliography style to be used, and
%% the bibliography file.
\bibliographystyle{ACM-Reference-Format}
\bibliography{main}

\newpage

\appendix

\section{Implementation Details}

We describe the implementation details for both offline and online experiments to support the reproducibility of our work.

\subsection{Feature Engineering of Offline Experiments}\label{app:feature_engineering}
\begin{itemize}[leftmargin=*]
\item \textbf{CACTRDC Dataset}: We extracted eight features from the raw features. Seven were discrete features, and the other was 100-equal-width discretized predicted click-through rate (pCTR). The features were selected according to our online feature importance evaluation system. We only adopted discretized features to facilitate not only the application of the tree structure, but also the engineering optimization of online performance (run-time and the consumption of computing resources), detailed in Section~\ref{add_online_deploy_impl}. 

\item \textbf{Porto Seguro and Avazu Datasets}: To make the experimental settings as consistent with CACTRDC, we only considered discretized features and selected seven discrete features based on GBDT's feature importance. We also took the 100-equal-width discretized predicted value as a calibration feature. For Porto Seguro, the selected seven features are ps\_ind\_05\_cat, ps\_ind\_17\_bin, and ps\_car\_01\_cat, ps\_car\_06\_cat, ps\_car\_07\_cat, and ps\_car\_09\_cat. For Avazu, the selected seven features are banner\_pos, site\_id, site\_category, device\_type, C16, C19, and C20.
\end{itemize}

\subsection{Online Deployment}\label{add_online_deploy_impl}
Our online model for CTR prediction is a DNN model and its AUC reaches about 0.77. The model uses several hundred features, mainly involving user profiles (e.g., age and area), user behaviors (e.g., recently clicked ads), position, current time, and properties of ad (e.g., text, image, and ID). 
Our offline CACTRDC experiment considered same set of features during feature selection.

In order to meet the stability and latency requirements of online service, we could transform the trained MBCT model into a systematic "if...then..." rules, which benefits a lot from the interpretability of tree model. Then, we used a high-speed cache to store these "rules" for online deployment. When new user-generated data were accumulated, we frequently re-trained and updated our MBCT model to ensure real-time calibration; this improved business profits for the online advertising platforms.

We also deployed the following engineering techniques for MBCT to accelerate the running time and reduce the memory overhead:

\textbf{Data aggregation}: The MBCT algorithm takes discrete binning features as input. If a feature is continuous, we converted it to discrete one, so the combinations of the features are finite and we can map the samples into finite groups. This means we can aggregate the samples with the same feature group and use the average statistics of one group as a new sample to replace the samples in this group. The data aggregation can largely reduce the use of computational resources.

\textbf{Parallelization}: We follow the parallelization method proposed by XGBoost~\cite{chen2016xgboost} for accelerating the running time of MBCT. In the process of constructing the calibration tree, we parallelize $s$ processes at any level of the tree for selecting the optimal binning feature, where $s$ is the number of features. This parallelization method guarantees load balancing to the largest extent, resulting in roughly $s$ times  run-time acceleration. Concretely, the time complexity is optimized from $O(k*\min(s,d)*s*r*n)$ to $O(k*\min(s,d)*r*n)$, where $k$ is the tree number,  $s$ is the number of features, $d$ is the max depth of the tree, $r$ is the hyper-parameter for MVCE, and $n$ is the number of train samples. 

\section{Additional Experimental Results}
\subsection{In-Depth Analysis}\label{app:more_in_depth_analysis}
We show additional analysis on the other two datasets, which are not presented in the main text due to space limit (cf., ~Section~\ref{in_depth_ana_main}).

\begin{figure}[!t]
  \centering
  \includegraphics[width=0.9\linewidth]{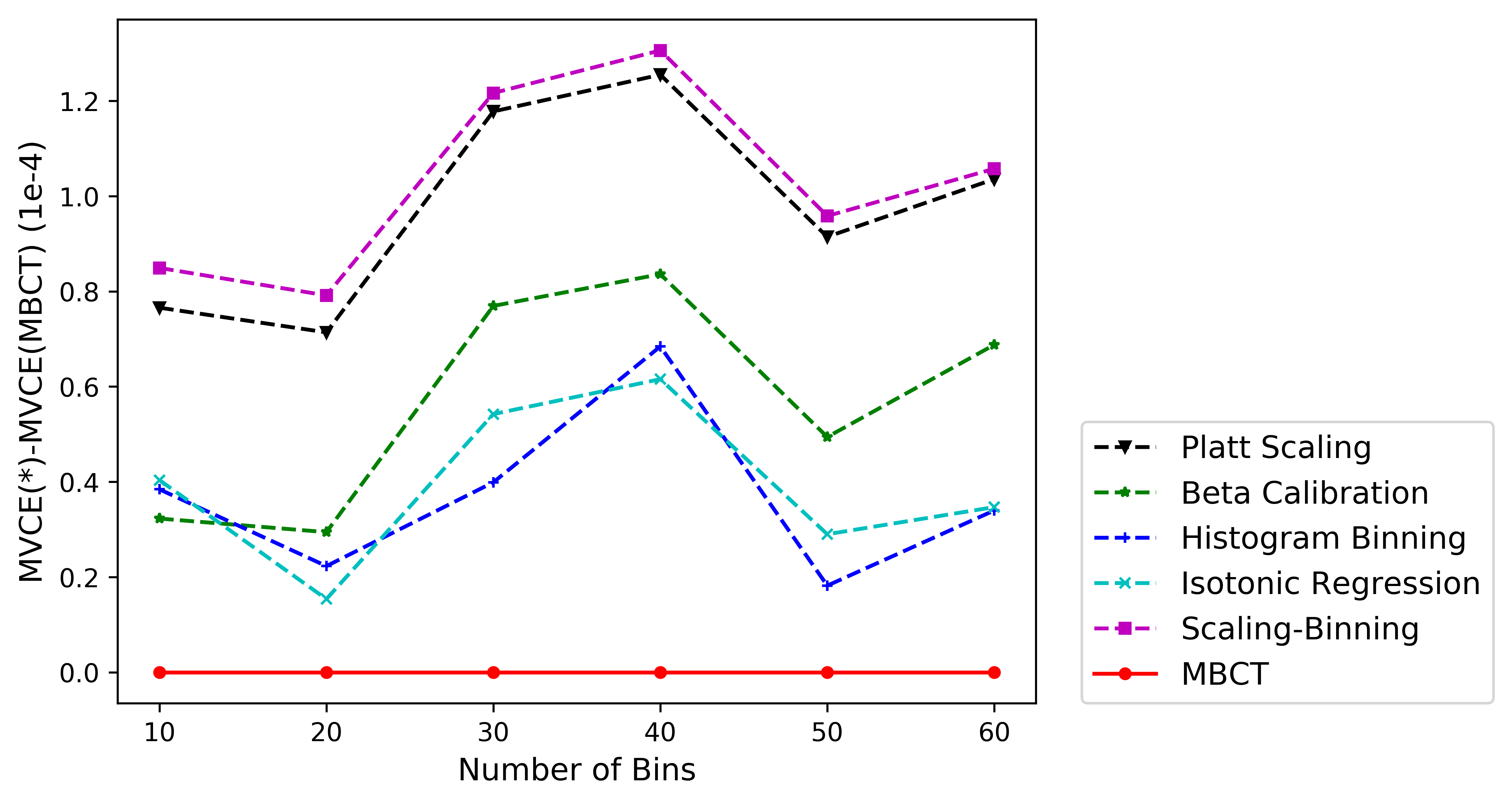}
  \caption{The difference of MVCE scores in comparison with our MBCT on Porto Seguro. Natually, our approach yields a value of 0. All competing models have a positive value, showing that our approach is the best. }\label{mvce_polyline_on_proto_seguro}
\end{figure}

\begin{figure}[!t]
  \centering
  \includegraphics[width=0.9\linewidth]{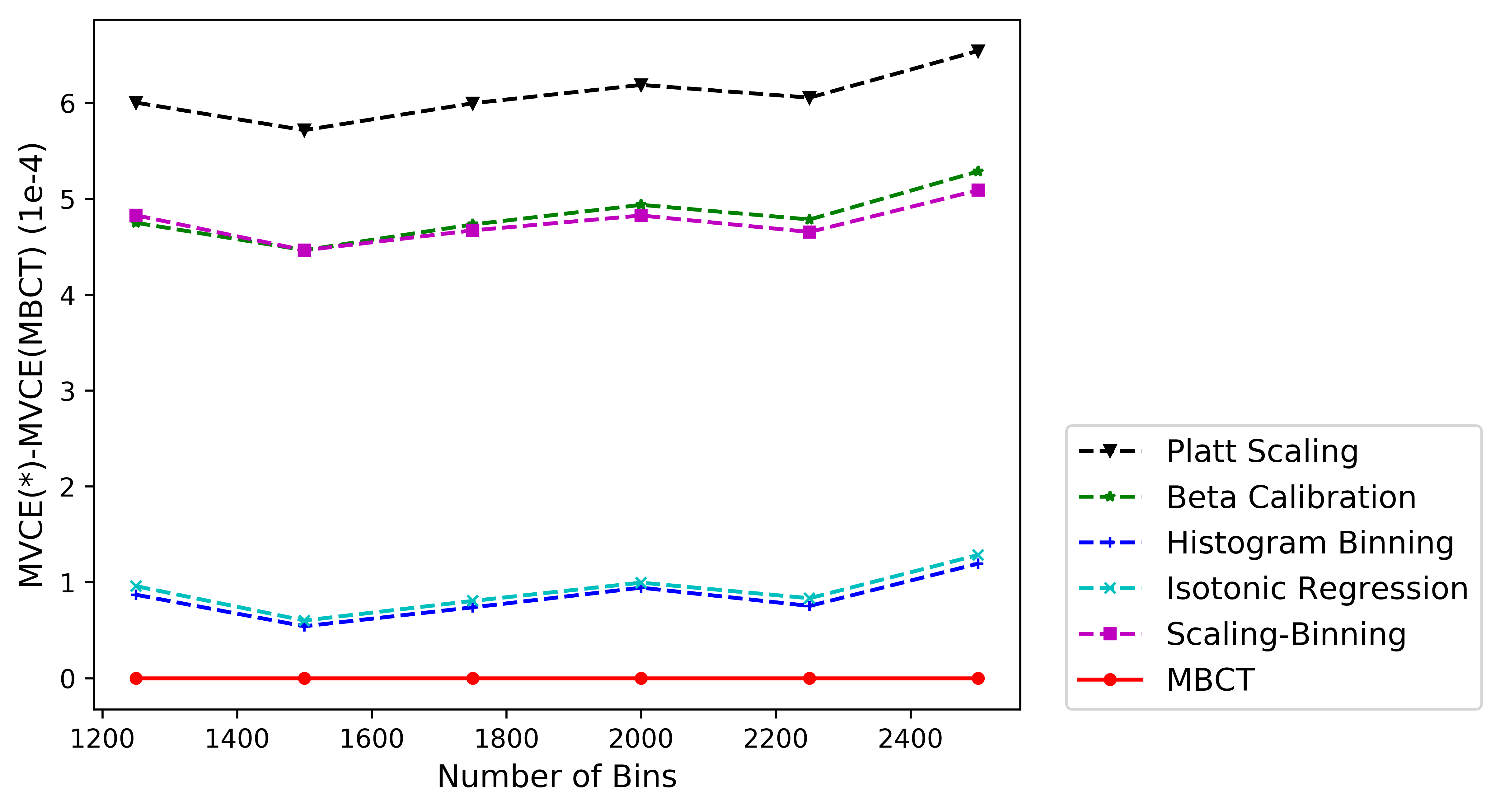}
  \caption{The difference of MVCE scores in comparison with our MBCT on Avazu.}\label{mvce_polyline_on_avazu}
\end{figure}

\begin{figure}[!t]
  \centering
  \includegraphics[width=0.95\linewidth]{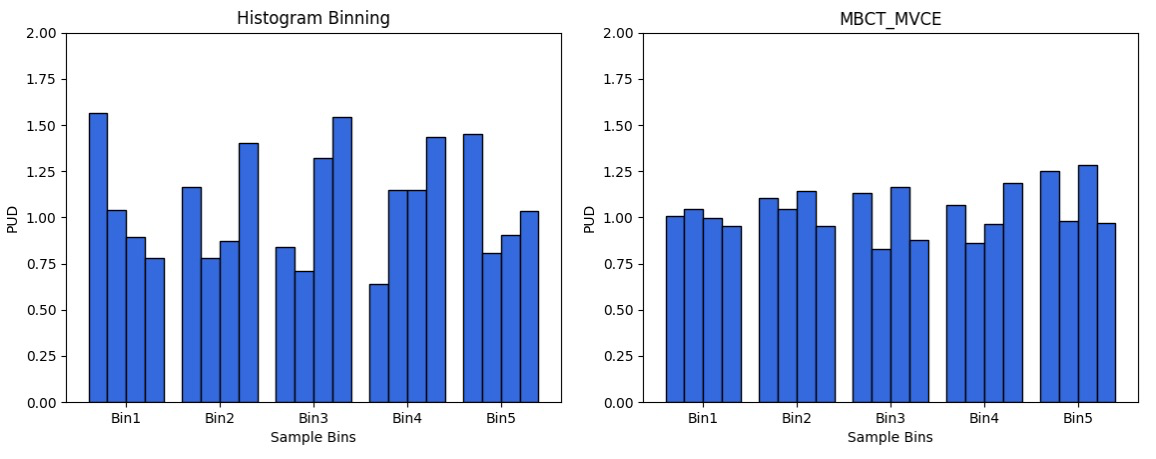}
  \caption{PUD of the finer-grained partitions in MBCT and Histogram Bining on Porto Seguro.}\label{case_study_on_proto_seguro}
\end{figure}

\begin{figure}[!t]
  \centering
  \includegraphics[width=0.95\linewidth]{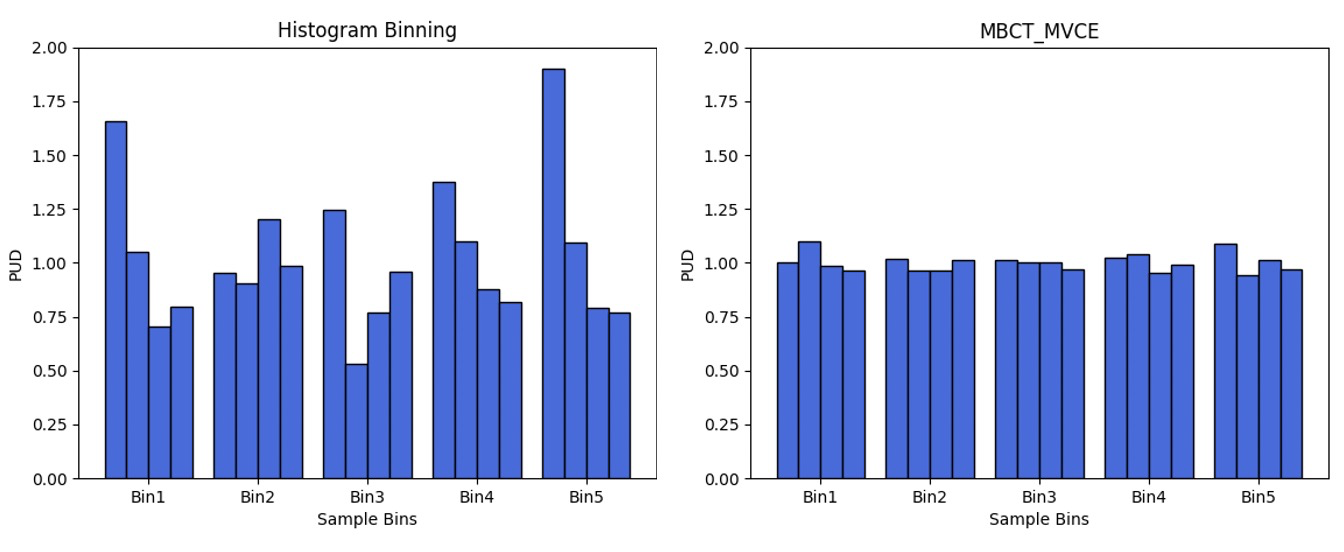}
  \caption{PUD of the finer-grained partitions in MBCT and Histogram Bining on Avazu.}\label{case_study_on_avazu}
\end{figure}

Figures~\ref{mvce_polyline_on_proto_seguro} and \ref{mvce_polyline_on_avazu} show the MVCE score under different bin sizes of the baselines and our method on Porto Seguro and Avazu, respectively. We see that MBCT also achieves the best MVCE score on these two public datasets, which is consistent with the results on real-world data of our online system. Figures~\ref{case_study_on_proto_seguro} and \ref{case_study_on_avazu} show the visualization of the PUD of finer-grained partitions on Porto Seguro and Avazu, verifying the effectiveness of feature-aware binning. All of the results in this section further illustrates that our method has a high generalization ability.

\subsection{Analysis of Calibration Metrics}\label{app:more_simulation}

In this part, we provide additional simulation experiments with different data distributions.

\begin{figure}[!t]
  \centering
  \includegraphics[width=0.90\linewidth]{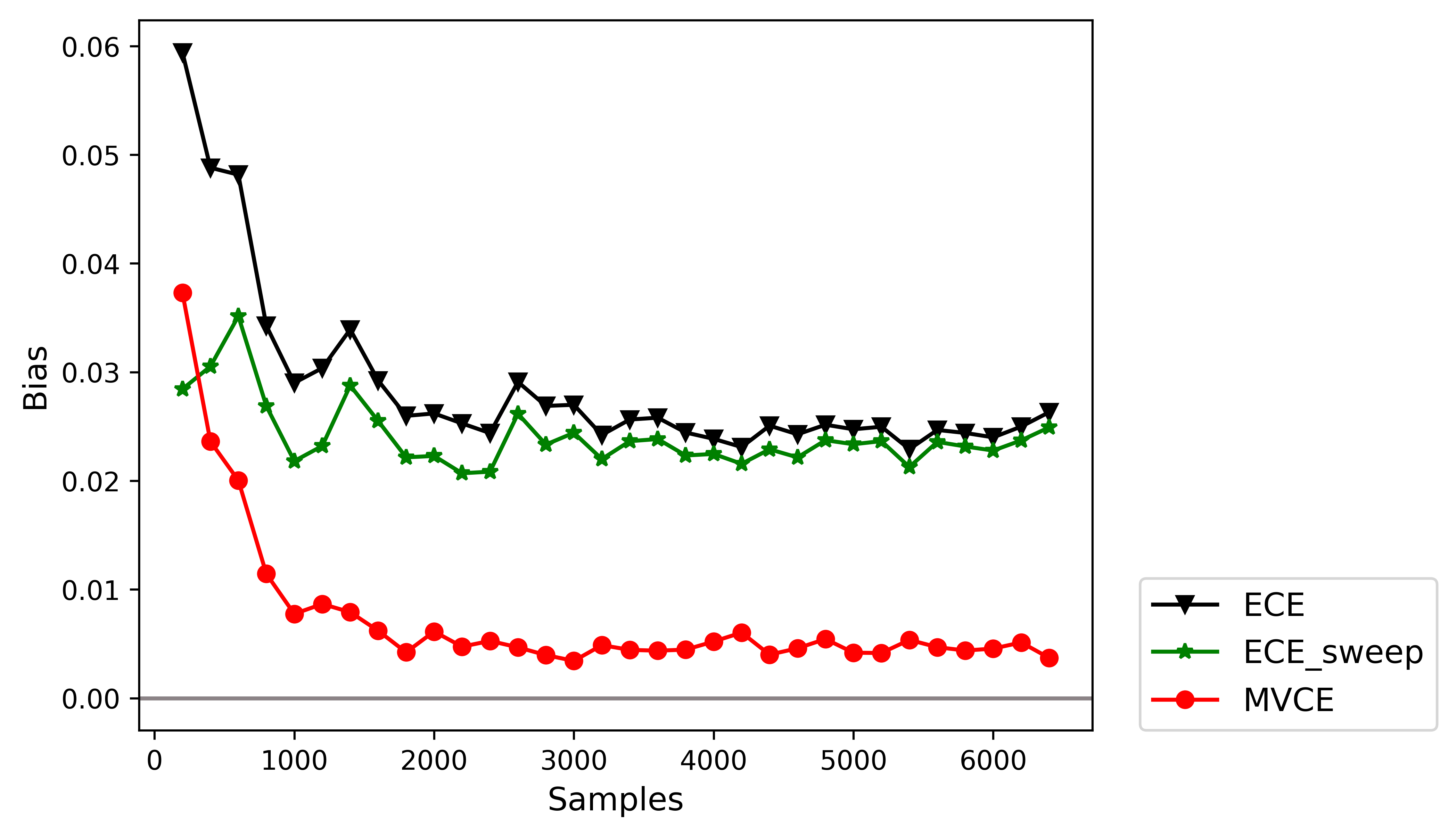}
  \caption{Main simulation results of ECE, $\text{ECE}_{\text{sweep}}$ and MVCE, in which $h(X)\sim \operatorname{Beta}(0.4,0.7)$ 
 and $\mathbb E[Y|h(X)=c]=c^2$. }\label{sim_polyline_04_07}
\end{figure}

\begin{figure}[!t]
  \centering
  \includegraphics[width=0.95\linewidth]{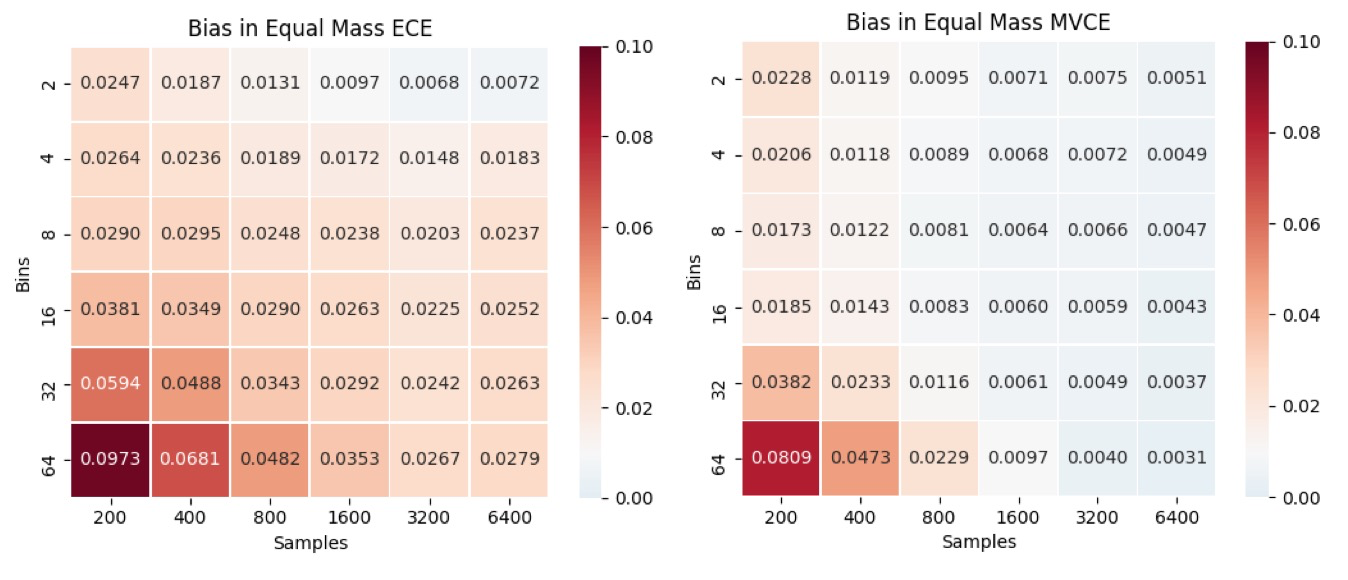}
  \caption{$\hat{E}_{bias}$ of MVCE and ECE under different numbers of bins and samples, in which $h(X)\sim \operatorname{Beta}(0.4,0.7)$ 
 and $\mathbb E[Y|h(X)=c]=c^2$. }\label{sim_heatmap_04_07}
\end{figure}

\begin{figure}[!t]
  \centering
  \includegraphics[width=0.90\linewidth]{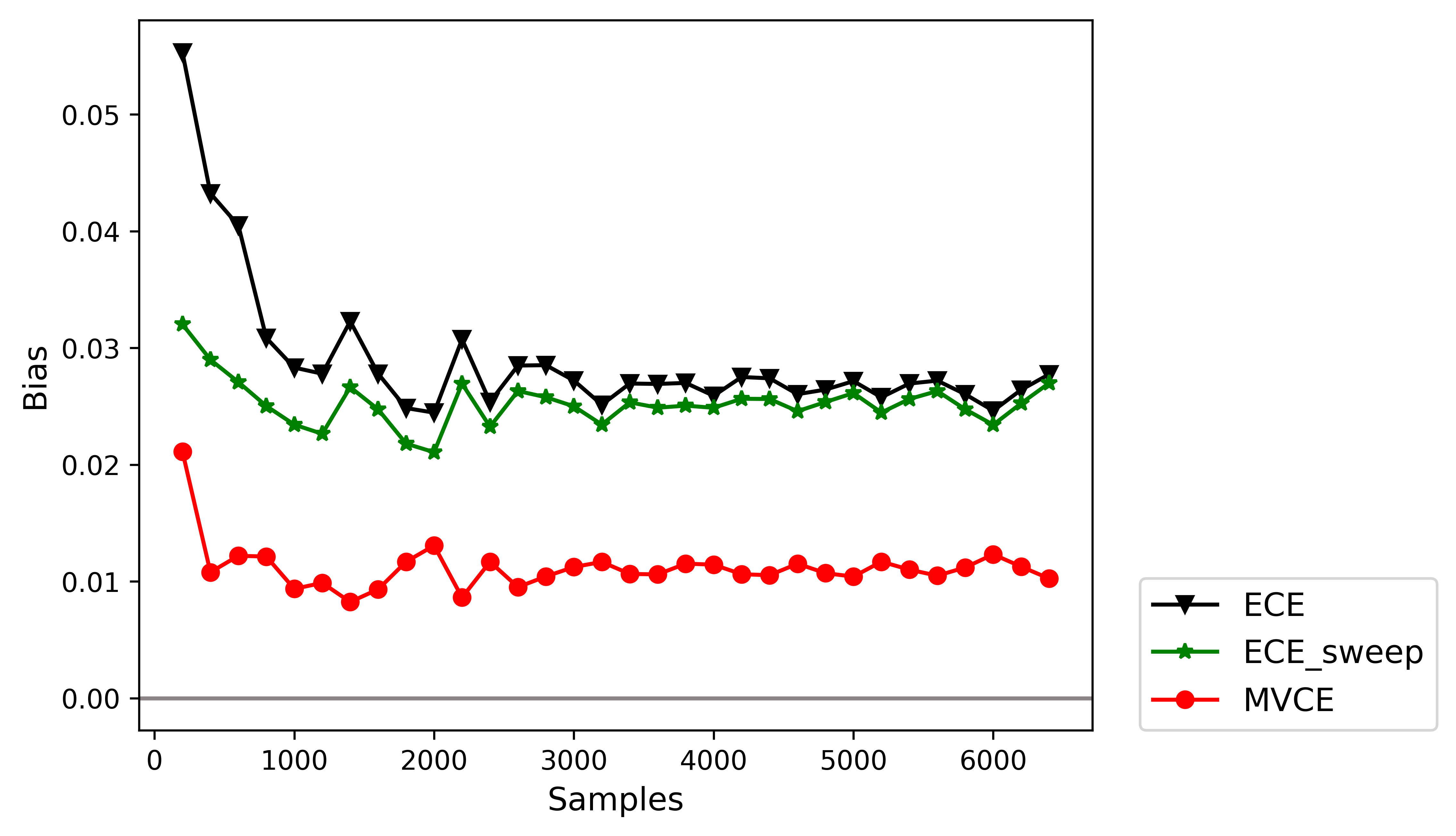}
  \caption{Main simulation results of ECE, $\text{ECE}_{\text{sweep}}$ and MVCE, in which $h(X)\sim \operatorname{Beta}(0.6,0.7)$ 
 and $\mathbb E[Y|h(X)=c]=c^3$.  }\label{sim_polyline_06_07_c3}
\end{figure}

\begin{figure}[!t]
  \centering
  \includegraphics[width=0.95\linewidth]{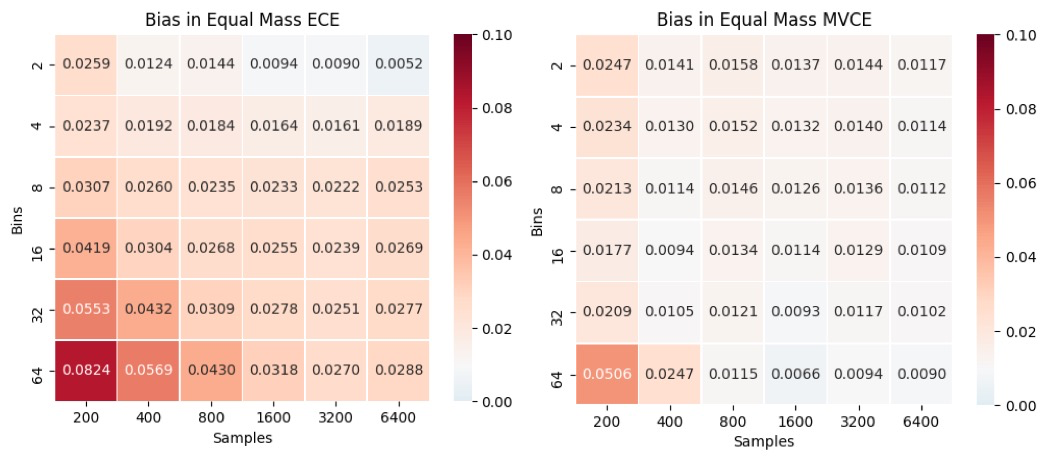}
  \caption{$\hat{E}_{bias}$ of MVCE and ECE under different numbers of bins and samples, in which $h(X)\sim \operatorname{Beta}(0.6,0.7)$ 
 and $\mathbb E[Y|h(X)=c]=c^3$. }\label{sim_heatmap_06_07_c3}
\end{figure}

Figures~\ref{sim_polyline_04_07} and \ref{sim_heatmap_04_07}
show the simulation results under $h(X)\sim \operatorname{Beta}(0.4,0.7)$ 
 and $\mathbb E[Y|h(X)=c]=c^2$.
Figure~\ref{sim_polyline_06_07_c3} and \ref{sim_heatmap_06_07_c3}
show the simulation results under $h(X)\sim \operatorname{Beta}(0.6,0.7)$ 
 and $\mathbb E[Y|h(X)=c]=c^3$. We see that the experimental results under these settings are consistent with the simulation experiment in the main text (in Section~\ref{sim_mvce}), which further indicates that MVCE is a more appropriate metric for calibration.
 
\section{Discussions of Calibration's Impact on Online Advertising Systems}
\label{app:discussion}
For online real-time bidding advertising platforms, the revenue of the platform is closely related to pCTR (predicted click-through rate). Take the simplest click-to-pay advertisement as an example, advertisers give their $bid$ on per click, the system allocate each chance of page-view to one advertiser and charge based on his $bid$ when the ad is clicked. The system should allocate the chance of a page-view to the advertiser whose ad achieves the highest score of $bid*pCTR$ to maximize its revenue. Thus, the order accuracy of $bid*pCTR$ score of the ads is crucial for the revenue of the platform. The $bid$ is given by advertisers, so the issue becomes to predict the true probability of the click-through rate. Therefore, calibration is quite important for real-time bidding advertising systems. The results of online experiments in the main text (Section~\ref{sec_online_exp}) provides empirical evidence that MBCT brings great business value to our online advertising system, even if the improvement of offline MVCE is seeming small (about 0.8\% relative improvement).

\end{document}